# The Term 'Agent' Has Been Diluted Beyond Utility and Requires Redefinition

**Brinnae Bent**

Duke University
brinnae.bent@duke.edu

**Abstract**

The term 'agent' in artificial intelligence has long carried multiple interpretations across different subfields. Recent developments in AI capabilities, particularly in large language model systems, have amplified this ambiguity, creating significant challenges in research communication, system evaluation and reproducibility, and policy development. This paper argues that the term 'agent' requires redefinition. Drawing from historical analysis and contemporary usage patterns, we propose a framework that defines clear minimum requirements for a system to be considered an agent while characterizing systems along a multidimensional spectrum of environmental interaction, learning and adaptation, autonomy, goal complexity, and temporal coherence. This approach provides precise vocabulary for system description while preserving the term's historically multifaceted nature. After examining potential counterarguments and implementation challenges, we provide specific recommendations for moving forward as a field, including suggestions for terminology standardization and framework adoption. The proposed approach offers practical tools for improving research clarity and reproducibility while supporting more effective policy development.

## Introduction

Since the advent of the term, what constitutes an "agent" in the field of artificial intelligence has been debated. The broad characterizations initially applied to the term were modified and altered to fit the needs of specific research disciplines, leaving a hodgepodge of definitions and no clear delineation of what can be deemed an "agent". With the wide adoption of large language model (LLM) systems, the term 'agent' has morphed once more, this time adopting a spectrum, where the terms 'agent' and 'agentic' - meaning agent-like - are used interchangeably.

This dilution of the term 'agent' in artificial intelligence is creating significant challenges in research communication, reproducibility, and system evaluation, which will ultimately cause challenges in policy development.

An example of the term's definitional multiplicity is highlighted by two recent papers presented at a prominent AI conference. Both papers center their work on the concept of "agents". In one paper, they describe agents as a hierarchical framework connecting different large language models together to make medical decisions (Kim et al. 2024).

In another paper exploring robotic control theory, they describe agents that are trained via reinforcement learning to complete open ended tasks like hide and seek and treasure finding within an environment (Zhang et al. 2024). While the methodology of the papers is markedly distinct from one another, the term 'agent' is core to their thesis - and the titles of their work. This presents two major problems. First, agent-based research is becoming increasingly difficult to evaluate as the inclusion criteria for 'agent' has evolved over time and system evaluations are not necessarily aligned with all definitions of the term. A number of recent benchmarks for AI agents evaluate proposed components of AI "agency" but fail to comprehensively define what constitutes an agent (Yao et al. 2024; Liu et al. 2024; Wu et al. 2024). The lack of standardization in the definition of AI agents results in multiple competing benchmark suites, each emphasizing aspects of AI agents based on their creators' implicit assumptions that may or may not overlap. No clear definition or standards for evaluation hinders the reproducibility of AI agent research.

The second problem hinges on the communication of AI agent research. Science communication is becoming increasingly important as public interest and use of AI-based tools becomes widespread; terminology disagreements will propagate beyond the research community to impact public perceptions and even policy development. While researchers may be able to compartmentalize different definitions for the term 'agent' in their work and that of others, it becomes increasingly difficult to communicate the nuances to those outside the field. For example, public-facing companies have different definitions of 'agent'. In a recent publication, Anthropic shares how they define agents, "Agents [...] are systems where LLMs dynamically direct their own processes and tool usage, maintaining control over how they accomplish tasks" (Anthropic 2024). Contrast this with the definition of agent from Langchain, "An AI agent is a system that uses an LLM to decide the control flow of an application" (Chase 2024). This lack of precision and standardization of the term seeps into communication to the general public. For example, the technology news firm VentureBeat recently shared this definition: "AI agents – software that can reason and perform specific business tasks using generative AI" (Marshall 2024). Leading voices in the field, such as Andrew Ng, have taken to social media to share their thoughts on the ambiguity of the term: "But there's a gray zone between what clearly is not an agent (prompting a model once) and what clearly is (say, an autonomous agent that, given high-level instructions, plans, uses tools, and carries out multiple, iterative steps of processing)" (Ng 2024).

The challenges in science communication that result from terminology uncertainty pose a greater threat than news articles: this communication is used to guide policy, specifically around AI system development, evaluation, and appropriate use. A nonprofit advocacy group recently defined "dimensions" of agent: "agenticity", which they borrow from OpenAI, and define as the "degree to which [the AI system] can adaptably achieve complex goals in complex environments with limited direct supervision" and "agentiality", which they define to be "the degree to which [the AI system] is actually authorized to ably represent the end user. A measure of relationship" (Whitt 2024). This definition has already been cited by those in public policy (Reisman and Whitt 2024). Without clear delineation of what constitutes an agent by the research community, policies may be developed using obsolete or incomplete definitions.

Given that terminology ambiguity impacts not only the evaluation of agent systems for research, but also science communication, public perception, and potentially public policy, we argue that there is a significant need to redefine what 'agent' means in the context of artificial intelligence. In this paper, we provide a historical analysis of the term, explore contemporary usage patterns, and use these as guiding principles for our proposed framework that defines clear minimum requirements for an AI system to be considered an agent and characterizes systems along a multidimensional spectrum. The primary objective being to provide more precise vocabulary for system description while preserving the term's historically multifaceted nature.

In this paper, we introduce the term agenticness to describe the degree to which a system exhibits characteristics commonly associated with an agent, such as environmental interaction, goal directed behavior, learning and adaptation, temporal coherence, and autonomy. We intentionally avoid the term agency, which carries substantial philosophical and sociological weight, referring to a being's capacity to act intentionally and meaningfully within a social or moral framework. Our objective is not to redefine agency in that sense, but to provide a practical vocabulary for evaluating AI systems that are frequently labeled as 'agents' in technical literature, regardless of whether they meet the deeper criteria of agency. The term agenticness is used here as a scoped, descriptive construct to discuss the presence and degree of agent-like attributes in artificial systems, without challenging the broader philosophical concept of agency.

## Early Definitions

The term 'agent' is not unique to the field of artificial intelligence. In chemistry, a chemical agent is a substance that can participate in or cause chemical reactions. Biological agents refer to organisms or substances that can affect living systems. In physics, physical agents are forms of energy or forces that can produce physical effects. Economic agents are decision-making entities in economic systems, and social agents are individuals or groups that can effect change in social systems. What makes the early definitions of the 'AI' agent distinct is its integration of multiple aspects of these earlier concepts: AI agents adapt to and influence their environment like chemical and biological agents. They make decisions based on available information like economic agents. And, like social agents, they have degrees of autonomy.

The term "agent" in AI originated early, likely borrowing terminology from adjacent fields like physics and biology. The first indexed publication containing the term "agent" was in the 1973 paper describing the actor model. They described ACTOR as "an active agent which plays a role on cue according to a script" (Hewitt, Bishop, and Steiger 1973). Agents were used to describe AI systems working symbiotically with humans (Chytil 1980) and multi-agent systems were described as existing in a single environment with goals and planning (McDaniel 1984). In the 1990s, we saw an explosion in the use of the term across the field of artificial intelligence (Wooldridge 1992; Vere and Bickmore 1990; Wavish and Graham 1995; Steels 1990; Singh 1991; Norman and Long 1995; Jennings 1995; Mitchell 1990; Franklin and Graesser 1996). A common vocabulary with formal definitions of agents for computer science, including robotic and software agents was developed in 1993, in which they describe general agent properties being successful (the extent that it accomplishes the specified task in the given environment), capable (possesses the effectors needed to accomplish the task), perceptive (distinguish characteristics of the world to achieve task), reactive (respond to events in the world), and reflexive (behaves in a stimulus-response fashion) (Goodwin 1995). They went on to describe "deliberative" agent properties as predictive, interpretive, rational, and sound (Goodwin 1995). Russell and Norvig defined an AI agent in their seminal textbook in 1995 as having the following characteristics: the ability to operate autonomously, perceive their environment, persist over a prolonged time period, adapt to change, and create and pursue goals. They further defined a "rational agent" as an agent that acts to achieve the best expected outcome (Russell and Norvig 1995). The first discussion of AI agenticness as a spectrum was introduced by Woolridge and Jennings, who described a "weak notion of agency" and a "stronger notion of agency", with the former having the properties of autonomy, social ability to interact with other agents (Genesereth and Ketchpel 1994), perception of their environment and the ability to respond in a timely fashion, and proactiveness, or exhibiting goal-directed behavior by taking initiative (Wooldridge and Jennings 1995). They attribute human-like properties to the "stronger notion of agency", including knowledge, belief, intention, and obligation (Wooldridge and Jennings 1995; Shoham 1993). Later texts attempted to simplify these definitions, defining an AI agent as taking sensory input from its environment and producing actions that affect its environment, where this interaction between the agent and its environment is usually ongoing and non-terminating (Weiss 1999).

The term 'agent' was widely adopted by the subfield reinforcement learning (Tan 1997; Claus and Boutilier 1998).

Sutton and Barton classify an agent in reinforcement learning as having goals relating to the state of the environment, being able to sense the state of the environment to some extent and having the ability to take actions that affect the state. They also argue that an agent should be able to learn from its own experience (Sutton and Barto 2015).

Even from the early days of the use of the term 'agent', there was a lack of consensus on the definition. A popular textbook from 1999 stated "Surprisingly, there is no such agreement: there is no universally accepted definition of the term agent, and indeed there is a good deal of ongoing debate and controversy on this very subject" (Weiss 1999). Similarly, Woolridge and Jennings described, "The problem is that although the term is widely used [...] it defies attempts to produce a single universally accepted definition" (Wooldridge and Jennings 1995). They acknowledged that this wasn't necessarily a problem but warned, "there is also the danger that unless the issue is discussed, agent might become a noise term, subject to both abuse and misuse, to the potential confusion of the research community" (Wooldridge and Jennings 1995).

Weiss posited that the only general consensus is that autonomy was central to the idea of a system being considered an agent (Weiss 1999). While most definitions included the fundamentals of autonomy, goal directed behavior, state awareness, and reactivity to their environment, there were many variations, with some definitions including mobility (White 1994); veracity (the agent will not knowingly communicate false information) (Galliers 1988), benevolence (Rosenschein and Genesereth 1985), or rationality (Galliers 1988; Rosenschein and Genesereth 1985).

## Contemporary Usage

While the term 'agent' is not new to the field of artificial intelligence, in recent years it has been applied in the context of large language model systems, further propagating the lack of consensus around its definition. The flexibility of large language models has led to their adoption across many applications that describe themselves as "AI agents". Early use of language model systems in the development of agent behaviors were in augmenting existing reinforcement learning experiments in either planning or language understanding, including the language component of an interactive SIMS-like experiment (Park et al. 2023), for the planning component of a goal-conditioned controller (Wang et al. 2024), and for teaching embodied agents (Wu et al. 2024). Recently work has incorporated large language model systems as the agents themselves, including for writing code to control skill-learning and game-playing simulations (Wang et al. 2023) and in combination with heuristics for text-based environments (Shinn et al. 2023).

Very recently, the term 'agent' has been used to describe a number of applications in which a pipeline consisting of large language models and either tools or external data sources is used to solve a task or set of tasks. Examples include a hierarchical framework connecting different large language models together to make medical decisions (Kim et al. 2024), a pipeline combining a large language model with tools including web searching, code execution, and experiment automation to conduct scientific research (Boiko, MacKnight, and Gomes 2023), and multiple role-playing language models prompted for communication purposes (Li et al. 2023). A recent study exploring whether agents can exhibit human trust behavior developed what they describe as an agent using a pipeline of language models and sophisticated prompt design (Xie et al. 2024).

Recent frameworks have converged on defining AI agenticness as a multidimensional spectrum rather than a binary property, though they emphasize different aspects of this continuum (Shavit et al. 2023; Chan et al. 2023; Kapoor et al. 2024). This spectrum is characterized by the degree to which systems can adaptably achieve complex goals in complex environments with minimal supervision (Shavit et al. 2023). Shavit et al. (2023) and Chan et al. (2023) both propose four-dimensional frameworks that share significant overlap: both emphasize the importance of autonomous goal achievement (termed "independent execution" and "directness of impact" respectively) and the ability to handle complex, long-term objectives (captured through "goal complexity" and "long-term planning") (Shavit et al. 2023; Chan et al. 2023). Kapoor et al. (2024) consolidates these dimensions into three broader categories, introducing explicit consideration of user interface and system design while maintaining focus on environmental complexity (Kapoor et al. 2024). These frameworks collectively suggest that agenticness emerges from the interaction between a system's autonomous capabilities, its ability to handle complexity across multiple timescales, and its capacity to operate with minimal human oversight.

The emergence of large language model systems has introduced new considerations to the conceptualization of the term AI agent, particularly through their potential as general problem solvers. While earlier frameworks focused on abstract properties of the term agent, recent work has begun to examine how these properties manifest in LLM-based systems (Kapoor et al. 2024; Weng 2023). Weng (2023) proposes a concrete architectural framework that implements many of the theoretical dimensions of agency: the system's planning capabilities address goal complexity and long-term planning, its memory systems enable environmental adaptation and goal persistence, and its tool integration facilitates direct real-world impact (Weng 2023). These definitions, by focusing primarily on large language models, overlook the diverse range of agent research being conducted in artificial intelligence.

Our work complements recent efforts such as Kasirzadeh and Gabriel (2025), which develop a governance-oriented framework for characterizing AI agents (Kasirzadeh and Gabriel 2025), and Sapkota et al. (2025), which distinguish between "AI Agents" and "Agentic AI" based on architectural paradigms and application domains (Sapkota, Roumeliotis, and Karkee 2025). While these frameworks emphasize policy implications and system taxonomies respectively, our goal is to offer a unifying evaluative structure grounded in operational properties in order to provide minimum criteria and graded dimensions applicable across research, benchmarking, and system design.

## Proposed Framework for Redefinition

Our framework brings together historical and contemporary perspectives on considerations to be an AI agent, acknowledging its roots in fields ranging from chemistry to economics while focusing on its evolution within the field of artificial intelligence. Building on Russell and Norvig's foundational characteristics of autonomy, environmental perception, persistence, adaptability, and goal-directed behavior, we propose a multidimensional analysis that reflects the capabilities of modern AI systems (Russell and Norvig 1995). Following recent frameworks by Shavit et al. and Chan et al., we recognize agenticness as existing on a spectrum rather than as a binary property, characterized by a system's ability to adaptably achieve complex goals with varying degrees of supervision (Shavit et al. 2023; Chan et al. 2023). While our framework builds on the multidimensional approaches of Shavit et al. (2023), which include goal complexity, environmental complexity, adaptability, and independent execution, we propose an expanded framework that introduces "temporal coherence" as a distinct dimension. Our framework considers how these properties manifest in large language model systems, which have introduced new considerations around general problem-solving capabilities and tool integration. We begin by establishing baseline requirements for a system to be considered an agent, then outline core dimensions that measure degrees of agenticness across different capabilities. Finally, we demonstrate how this framework can be applied to categorize and evaluate contemporary AI systems, from simple automated tasks to advanced multi-model pipelines that combine planning, tool use, and environmental interaction.

## Minimum Requirements to be Considered an Agent

For a system to qualify as an agent, it must meet three basic criteria. These criteria were derived from recurring themes in the literature (Russell and Norvig 1995; Shavit et al. 2023; Chan et al. 2023).

To be considered an agent, the system must have an active and measurable environmental impact. The system should be capable of taking actions that meaningfully and persistently alter its environment. These changes should be observable and/or measurable by other parts of the system or external observers. Simple input-output mappings (e.g., basic chatbots) do not qualify as agentic under this requirement.

To be considered an agent, the system must have goal-directed behavior. The system should operate in service of defined objectives, whether externally specified or internally generated. Goals should be adaptive based on the environment rather than having fixed optimization targets. The system must pursue objectives through multi-step planning and execution, not just optimization. Purely reactive systems (e.g., a language model that selects responses based only on immediate context without maintaining longer-term conversational goals or developing multi-step strategies to achieve user objectives) do not qualify as agents.

To be considered an agent, the system must have state awareness. The system should maintain and update some representation of the environmental state that influences its decisions. This state awareness should persist between interactions. Stateless systems (e.g., a model that processes each input independently without maintaining context or memory) do not qualify as agents.

## Agent Core Dimensions

Once an AI system has met the minimum requirements to be defined as an 'agent', they can be placed along a spectrum of five key dimensions, inspired by recent definitions of the agentic spectrum (Shavit et al. 2023; Chan et al. 2023; Kapoor et al. 2024). These dimensions provide a framework for evaluating and comparing different AI systems' levels of agenticness, from threshold capabilities to highly complex implementations (Figure 1).

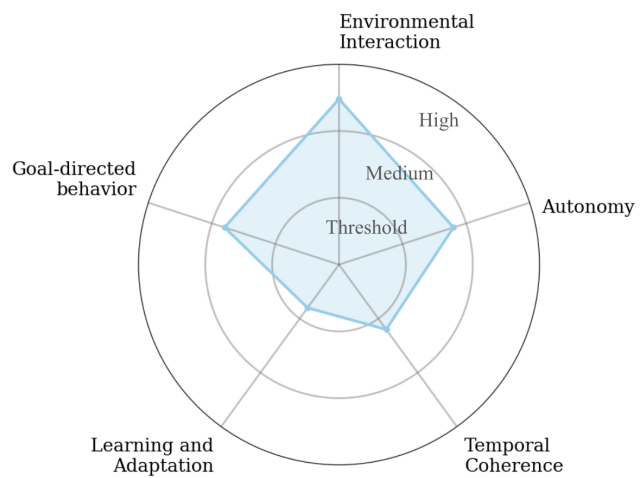

Figure 1: The core dimensions of agenticness. The system described by this visualization exhibits threshold level learning and adaptation (inner circle), intermediate level temporal coherence, goal-directed behavior, and autonomy (middle circle) and advanced level environmental interaction (outer circle).

**Environmental Interaction Sophistication** Environmental interaction sophistication represents an AI system's ability to perceive, understand, and manipulate its operational environment, whether physical, digital, or abstract. We include abstract environments in this definition to future-proof this definition, as environments may not involve physical or digital spaces, such as in mathematical theorem spaces or ethical frameworks.

Our primary focus on differentiating levels of environmental interaction sophistication focusses on action spaces, which are categorized by their flexibility and complexity. Predefined action spaces restrict systems to fixed, pre-programmed actions in structured environments, like a manufacturing robot following set routines. Flexible action se-lection enables systems to choose from a wider range of possibilities based on context, adapt to semi-structured environments, and demonstrate basic tool use, as seen in autonomous delivery robots. Complex action composition allows systems to create novel combinations of actions for unfamiliar problems, operate in unstructured environments, and show sophisticated tool use by appropriately combining tools and techniques for multi-step tasks.

At the threshold level, systems demonstrate basic environment modeling with a defined action space. For example, a robotic arm in a manufacturing setting that can identify and pick up specific parts from predefined locations demonstrates threshold agenticness. The system understands its immediate environment and the direct consequences of its actions.

Intermediate level systems develop multiple input modalities and flexible action selection. Intermediate-level systems have basic tool use capabilities. A digital marketing AI that processes user behavior data, adjusts campaign parameters, and coordinates across multiple platforms exemplifies this level in abstract environments. Similarly, an autonomous delivery robot navigating through semi-structured physical environments demonstrates comparable sophistication in the physical realm. It can process visual, spatial, and sensor data while adapting its path based on obstacles and changing conditions.

Advanced-level systems exhibit multi-modal perception and complex action composition in unstructured environments. An example of an advanced-level system is a household robot that can recognize various objects, understand verbal instructions, and perform complex tasks like cooking or cleaning. It demonstrates sophisticated tool use by selecting and combining different kitchen utensils appropriately for each recipe.

**Goal Directed Behavior Complexity** Goal complexity and management reflects a system's ability to form, understand, and pursue objectives that adapt to environmental conditions while engaging in multi-step planning and execution.

Threshold systems demonstrate basic adaptive goal-directed behavior within a narrow domain. Consider an inventory management system that modifies a primary objective based on immediate environmental conditions. While maintaining its core goal of efficient inventory management, it can adapt its approach based on factors like current storage capacity or urgent orders. The system develops simple multi-step strategies, such as reorganizing a section of the warehouse in response to changing inventory levels, showing basic goal-directed behavior beyond mere optimization.

Intermediate level systems coordinate multiple interdependent goals across different domains. An AI project management assistant exemplifies this level by managing interconnected objectives across team performance, resource allocation, and project timelines. It recognizes how adjusting one goal (such as accelerating a deadline) impacts others (like resource requirements and team workload) and can proactively modify its objectives across these dimensions. The system develops advanced strategies that account for complex dependencies between goals, demonstrating a higher level of planning capability than threshold systems.

Advanced level systems demonstrate abstract goal formation and complex adaptive planning. An AI education system represents this level by developing and modifying personalized learning objectives based on student performance, engagement patterns, and emerging learning opportunities. It can form abstract goals like "develop critical thinking skills" while creating concrete, adaptive strategies to achieve them. The system continuously refines both its objectives and approaches based on student progress, learning style, and changing educational requirements, demonstrating goal-directed behavior that goes well beyond simple optimization or reactive responses.

**Temporal Coherence** Temporal coherence represents a system's ability to maintain consistent operation over time through state awareness and memory. Temporal coherence is a foundational capacity that enables and constrains goal-directed behavior and environmental interaction. However, it is conceptually distinct from both, as it governs the consistency and continuity of state representations over time, which is needed for long-horizon tasks and multi-turn reasoning. Unlike static expert systems or declarative knowledge bases, temporal coherence encompasses not just stored knowledge, but the active maintenance, update, and integration of memory representations across time. This dynamic function is essential to adaptive, agentic behavior.

Threshold systems show basic state maintenance and short-term memory. A smart home system that maintains temperature settings and remembers recent user preferences demonstrates this level. It can maintain an immediate state but has limited historical awareness.

Intermediate-level systems develop persistent state modeling and working memory systems. An AI personal assistant that maintains context across conversations and remembers user preferences over time in order to curate an event calendar, travel plans, and make calls on the user's behalf exemplifies this level. It can track ongoing tasks and adapt to evolving user needs.

Advanced-level systems exhibit advanced state management through hierarchical memory structures and complex temporal reasoning. Consider an autonomous scientific research system that maintains multiple layers of state representation: it tracks immediate experimental conditions, maintains medium-term project trajectories, and builds long-term understanding of research domains. The system can integrate new findings with existing knowledge, recognize patterns across different time scales, and adjust its research strategies based on both historical insights and new ideas. For example, when investigating a new phenomenon, it can reference failed approaches from months ago, connect them with recent successful methodologies, and synthesize this information to develop novel experimental designs. This demonstrates advanced state management by maintaining coherent operation across multiple time horizons while actively using temporal patterns to inform decision-making.

**Learning and Adaptation** The learning and adaptation dimension encompasses a system's capacity to improve its performance and adjust to new situations over time.

Threshold systems show basic parameter updating and learning within defined scenarios. Consider a document classification system that updates its classification thresholds through periodic batch training on validated data sets. It can improve its performance through supervised learning in controlled training sessions but cannot adapt its parameters in real-time during operation. The system learns within its predetermined domain through discrete update cycles rather than continuous adaptation.

Intermediate-level systems feature online learning capabilities and adaptation to common variations. An industrial quality control system that can adapt to new product variations and environmental conditions exemplifies this level. It accumulates knowledge about different defect types and can generalize to similar cases.

Advanced-level systems demonstrate continuous learning, knowledge synthesis, and meta-learning capabilities. Consider an adaptive traffic management system that orchestrates multiple urban districts. This system continuously learns from real-time traffic patterns, weather conditions, and special events while synthesizing historical data to improve its control strategies. When encountering new situations, such as previously unseen combinations of weather events and traffic patterns, it can develop new management approaches by combining and adapting existing strategies. The system demonstrates sophisticated learning by maintaining multiple specialized models for different traffic scenarios (e.g., rush hour, special events, emergency response) and can intelligently select and combine these models based on current conditions. It can also transfer knowledge between different urban areas, adapting successful strategies from one district to another while accounting for local variations in road layout and typical usage patterns. This learning behavior is advanced because of its ability to continuously improve, combine knowledge across domains, and adapt its modeling approach based on context.

**Autonomy** Autonomy characterizes a system's ability to operate without constant external guidance, including its capacity to handle errors and unexpected situations. This definition is informed by Müller's (2012) conception of autonomy as a relational and graded property, where an agent is autonomous to the extent that it can act without guidance from others (Müller 2012). Froese et al. (2007) further distinguish between behavioral autonomy, or flexible goal-directed behavior in dynamic environments, and constitutive autonomy, the system's ability to self-organize and maintain its own identity (Froese, Virgo, and Izquierdo 2007). Dignum (2018) emphasizes that increasing autonomy must be accompanied by increasing ethical responsibility, highlighting the need for autonomous systems to align with human values and remain accountable (Dignum 2018). Thus, autonomy is not only a spectrum of technical capability, but also as an ethically grounded and organizationally layered construct.

Where adaptation primarily concerns a system's capacity to improve performance and adjust to new situations over time through learning, parameter updating, and knowledge accumulation, autonomy refers to a system's ability to operate independently without constant external guidance, with particular emphasis on error handling, recovery from unexpected situations, and self-directed operation. While both dimensions address handling unexpected situations, adaptation is forward-looking, learning from experiences to enhance future performance, whereas autonomy is immediate, maintaining effective operation when challenges arise.

Threshold systems demonstrate bounded autonomous operation with basic error handling capabilities. Consider an automated trading system that operates independently within strictly defined parameters. When encountering known error conditions, such as unusual market volatility or data inconsistencies, it can execute predetermined fallback strategies. The system maintains operational safety through clear decision criteria and basic error recovery procedures, such as halting trades or reverting to conservative trading limits when it detects anomalies in market data.

Intermediate-level systems achieve extended autonomous operation with robust error management. An autonomous drone performing site inspections exemplifies this level through its ability to handle various unexpected situations. It can independently navigate different environments while managing multiple types of errors: adjusting its flight path for weather conditions, rerouting around unexpected obstacles, and implementing alternative inspection procedures when primary sensors malfunction. The system can diagnose problems and select appropriate recovery strategies from a range of options, demonstrating flexible decision-making in both normal operations and error scenarios.

Advanced-level systems exhibit self-directed operation with clever error handling and recovery capabilities. Consider an AI manufacturing control system that coordinates multiple production lines while managing complex interdependencies. When confronted with novel error conditions, it can analyze the situation, develop new solutions, and implement recovery strategies that consider both immediate impacts and long-term consequences. For example, if it encounters an unprecedented combination of equipment failures, it can reorganize production workflows, reallocate resources, and develop new quality control procedures to maintain operations. The system demonstrates advanced problem-solving by generalizing from past experiences to handle novel errors, while maintaining operational integrity across multiple subsystems.

### Dimensional Interdependencies

The five core dimensions of agenticness exist not as isolated characteristics but as interconnected aspects that collectively define an AI system's capabilities. While each dimension can be evaluated independently, there exists interaction effects which impact the spectrum.

For example, a system's ability to maintain state awareness directly impacts its environmental interaction capabilities. Higher temporal coherence enables more complex environmental modeling and prediction, while more challenging environmental interactions often require more advanced state management. Similarly, a system's learning capabilities significantly influence its ability to form and pursue complex goals. As learning sophistication increases, the system can develop more granular goal hierarchies and better adapt its objectives based on environmental feedback. Conversely, more complex goal structures often drive the development of more advanced learning mechanisms. Temporal coherence and learning adaptation have high interdependence. A system's ability to learn effectively often depends on its capacity to maintain and utilize historical information, while improved learning capabilities can enhance the system's ability to develop more innovative temporal models. The level of autonomy a system can achieve is closely tied to its ability to understand and manipulate its environment. More sophisticated environmental interaction capabilities enable higher degrees of autonomous operation, while increased autonomy often drives the development of more advanced environmental interaction mechanisms.

## Application of the Proposed Framework

Several common classes of AI systems fail to meet the minimum requirements to be considered an agent outlined in our framework. Simple chatbots, while capable of generating contextually appropriate responses, typically lack meaningful environmental impact and maintain no persistent state representation. Their behavior, though impressive in terms of language generation, remains fundamentally reactive rather than goal-directed. Basic classification systems represent another category that falls below the threshold. Despite potential complexity in their pattern recognition capabilities, these systems operate through pure input-output mapping without environmental intervention or goal-directed behavior. Their operation, while valuable for specific tasks, does not demonstrate the key characteristics required for being considered an agent under our minimum requirements. Static expert systems, despite their historical significance in AI development, similarly fail to qualify as agents. Their lack of learning capabilities and reliance on predefined responses places them below the minimum threshold to be considered an agent. While these systems can make complex decisions within their domain, they lack the essential characteristics of environmental interaction and adaptation that constitute being an agent.

To illustrate how our framework applies to more advanced systems, we consider four example classes of AI agents (Figure 2) and qualitatively assess them based on our framework. Note that this is purely illustrative and based on documented (or theoretical) system capabilities. Future research into evaluation and benchmarking of each core dimension would enable quantitative assessment in addition to qualitative evaluation of dimensions:

The Smallville generative agents demonstrate sophisticated levels of agenticness across multiple dimensions (Park et al. 2023) (Figure 2a). They exhibit intermediate to advanced level environmental interaction through their ability to navigate physical spaces, interact with objects, coordinate with other agents, and maintain representations of their environment through subgraph memories, though they operate in structured sandbox environment. They exhibit intermediate-level goal-directed behavior as they can form and pursue multiple objectives through planning and coordination, adapting to environmental changes and other agents' actions. They exhibit intermediate to advanced level temporal coherence through their memory stream system that maintains comprehensive records of experiences and enables relevant memory retrieval for decision-making, though challenges with long-term coherence remain even with advanced models. Their learning and adaptation capabilities sit at the intermediate level, as they can update their understanding of the environment and relationships over time, though the paper doesn't specify advanced learning mechanisms beyond memory accumulation and retrieval. Autonomy reaches intermediate to advanced levels as they can operate independently, handle various social interactions, and navigate complex environments while maintaining behavioral coherence, though they still operate within the constraints of their sandbox world and language model capabilities.

LLM-based personal assistants are becoming popular (Figure 2b). An AI personal assistant that maintains context across conversations and remembers user preferences over time in order to curate an event calendar, travel plans, and make calls on the user's behalf demonstrates varying levels of agenticness across different dimensions. It exhibits intermediate level environmental interaction through its ability to interact with multiple digital systems and platforms, handling various input modalities and showing basic tool use capabilities in scheduling, travel booking, and communication, though it lacks sophisticated action composition in unstructured environments. In goal-directed behavior, it reaches the intermediate level by coordinating multiple interdependent goals across scheduling, travel planning, and communication while managing competing priorities and constraints, although its goals remain primarily user-defined rather than self-generated. The system exhibits intermediate level temporal coherence through its ability to maintain persistent state modeling across conversations, remember user preferences, and track ongoing tasks, though it lacks more sophisticated hierarchical memory structures. In learning and adaptation, it remains at the threshold level, as it can update user preferences based on feedback but shows limited online learning capabilities and operates primarily within predefined parameters without demonstrating knowledge synthesis or meta-learning. Autonomy is at the intermediate level, reflecting its ability to operate independently for routine tasks and handle basic error conditions, while still requiring user confirmation for significant decisions and showing limitations in managing novel or complex situations.

A standard vacuum cleaning robot demonstrates varying levels of agenticness across different dimensions (Figure 2c). In terms of environmental interaction, it falls into the threshold category due to its basic sensory suite of cameras and bump sensors, which enable fundamental environment modeling and obstacle avoidance, though it lacks sophisticated manipulation capabilities. Its goal-directed behavior also sits at the threshold level, as it primarily focuses on a single cleaning objective with basic adaptive behaviors and simple multi-step strategies like navigation and recharging, but without managing multiple interdependent goals.

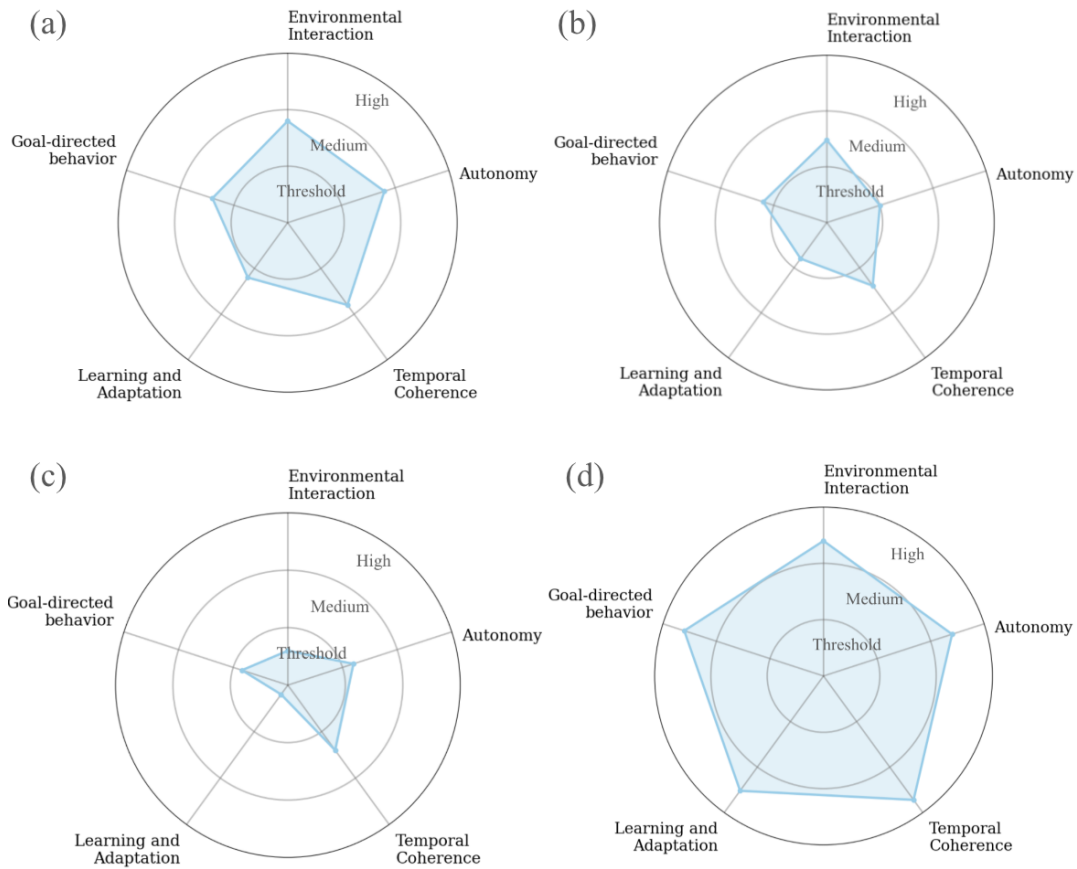

Figure 2: Examples of core dimensions for (a) Smallville generative agents (Park et al. 2023), (b) LLM-based personal assistant, (c) Standard vacuum cleaning robot, (d) Theoretical autonomous scientific research system

The robot exhibits intermediate level temporal coherence, demonstrated by its ability to maintain persistent state information through map storage, cleaning history tracking, and charging station memory, though it doesn't exhibit more complex temporal reasoning. Learning and Adaptation is its weakest dimension, as it operates with only basic parameters and cannot improve its performance over time through experience or develop new strategies. The robot exhibits intermediate level autonomy, having extended independent operation with the ability to handle basic error conditions and navigate various environments while maintaining fundamental recovery strategies like returning to its charging station when battery is low.

A theoretical autonomous scientific research system demonstrates exceptionally high levels of agenticness across all dimensions (Figure 2d). It exhibits advanced level environmental interaction through its ability to interact with complex experimental setups, manipulate multiple variables and tools, and maintain perception of experimental conditions and outcomes. It also exhibits advanced level goal-directed behavior, as it manages multiple interconnected research objectives, forms abstract goals around scientific discovery, and develops complex, adaptive experimental strategies based on findings. The system excels particularly in temporal coherence, featuring a sophisticated multi-layer state representation that integrates information across different time scales, from immediate experimental conditions to long-term research trajectories, while demonstrating advanced pattern recognition and historical reasoning capabilities. In learning and adaptation, it achieves the advanced level through its continuous synthesis of new findings with existing knowledge, sophisticated knowledge transfer between experiments, and meta-learning capabilities in experimental design. In autonomy, it also reaches the advanced level, as it operates independently in complex research scenarios, makes sophisticated decisions about experimental directions, handles unexpected results effectively, and self-directs research trajectories based on its findings.

The classification of certain AI systems presents particular challenges, highlighting important boundaries in our framework. Large Language Models (LLMs) with tool access represent a significant edge case. When equipped with capabilities to interact with external systems, these models may cross the threshold into being considered an agent if they demonstrate genuine environmental impact through their tool use, maintain clear state representations, and exhibit goal-directed behavior beyond simple pattern matching. However, this classification depends heavily on the specific implementation and capabilities of the tool integration system. Recommendation systems present another interesting boundary case. These systems may qualify as agents when they maintain persistent user state, demonstrate clear goal-directed behavior in optimizing recommendations, and produce meaningful environmental impact through user interaction patterns. However, many recommendation systems remain below the threshold, operating primarily through statistical pattern matching without true goal-directed behavior or state awareness.

This analysis of non-agent systems and edge cases yields several important implications for the application of our framework. First, the distinction between pattern recognition and true agenticness emerges as crucial. Systems may demonstrate impressive capabilities in specific domains while still failing to qualify as agents under our framework. Second, the interaction between different aspects of agenticness becomes apparent in edge cases, where systems may meet some but not all minimum requirements. The framework also highlights the importance of implementation details in determining what constitutes an agent. Two systems with similar high-level architectures may fall on different sides of the threshold based on specific details of their implementation and deployment. This observation underscores the need for careful evaluation of system capabilities rather than reliance on architectural classification alone.

## Discussion

### Alternative Views

Redefining any concept, no matter how ambiguous, is challenging. In this section, we address primary counterarguments to our proposed framework.

A significant counterargument to our framework comes from researchers who advocate for maintaining broader, more inclusive definitions of what should be considered an AI agent (Ng 2024). Attempting to impose stricter definitional boundaries might stifle creativity and exclude valuable research directions. While we acknowledge that definitional flexibility can foster innovation by allowing researchers to explore diverse approaches under shared conceptual frameworks, we contend that the current breadth of the term 'agent' has reached a point where it hinders rather than helps scientific discourse. Our framework preserves flexibility while adding necessary precision through its dimensional approach.

The risk of oversimplifying complex AI systems through dimensional analysis represents another potential concern. It could be argued that reducing the definition of an AI agent to five dimensions fails to capture the full complexity of modern AI systems and their capabilities. This concern particularly applies to emerging technologies that might exhibit novel forms of agenticness not easily captured by our current dimensional framework. This criticism warrants careful consideration, as oversimplification could indeed lead to misunderstanding or mischaracterization of AI systems. However, our framework addresses this concern in several ways. First, the dimensional approach explicitly acknowledges the complexity of agenticness by refusing to reduce it to a binary property. Second, the framework remains extensible, allowing for the addition of new dimensions as our understanding of AI capabilities evolves.

Another significant counterargument is in the practicality of implementing and enforcing new definitional standards in an already complex field. Attempting to impose new terminology on an established research community could be considered impractical and potentially counterproductive. This concern is particularly relevant given the decentralized nature of AI research and the rapid pace of technological development. We acknowledge these implementation challenges. Rather than attempting to enforce strict definitional compliance, our approach provides a useful tool that researchers can adopt as it proves valuable to their work. The framework's dimensional analysis offers immediate practical benefits for research communication and system evaluation, creating natural incentives for adoption.

For over 50 years, the definition of 'agent' in the field of artificial intelligence has remained ambiguous. A practical counterargument concerns the institutional and professional inertia that might resist terminology change. Changing established practices requires substantial effort and resources. This resistance to change could limit the framework's adoption regardless of its merits. We posit that the documented costs of terminological ambiguity in AI agent research and science communication outweigh the challenges of implementing a more precise framework. These challenges, while significant, ultimately highlight the need for rather than invalidate our approach. The framework provides a structured way to define AI agents while remaining adaptable to emerging developments in the field.

### Recommendations for Action

To move forward effectively, we propose the following recommendations for the adoption of precise and standardized definitions of agent terminology. We urge researchers to specify which dimensions of agenticness they are addressing and to what degree in their publications and communications. We also encourage benchmark developers to explicitly state which dimensions of agenticness their evaluations measure. This clarity will help researchers select appropriate evaluation metrics and understand the limitations of their results. Additionally, new benchmarks should be developed to specifically assess different dimensions of agenticness. We recommend that standards organizations and regulatory bodies adopt more precise terminology when developing AI governance frameworks. The dimensional approach provides a foundation for more granular policy that can adapt to technological development while maintaining consistent principles.

### Future Research Directions

Our framework for AI agents opens several avenues for future investigation. The field requires robust quantitative metrics to evaluate agent thresholds consistently across different types of AI systems. While our framework provides qualitative criteria, developing measurable standards is needed to classify and compare AI systems effectively.

The phenomenon of emergent agenticness demands particular attention as AI systems evolve. Current systems often operate at the boundaries of our framework, exhibiting characteristics that challenge traditional definitions of what it means to be considered an agent. Studying these boundary cases could reveal fundamental insights about how agenticness manifests in artificial systems and how it might evolve as technology advances.

These research directions carry significant ethical implications. The development of increasingly agent-like AI systems raises fundamental questions about responsibility and autonomy. Who bears responsibility for the actions of highly agentic systems? How do we establish appropriate boundaries for artificial agents? As AI capabilities continue to expand, addressing these ethical considerations becomes increasingly urgent for both researchers and practitioners in the field.

## Conclusions

The dilution of the term 'agent' in artificial intelligence is creating significant challenges for research evaluation, reproducibility, communication, and policy development. While the term's flexibility has historically fostered innovation, its current state of ambiguity threatens to impede scientific progress and effective governance. The proposed framework addresses these challenges by establishing clear minimum requirements for a system to be considered an agent while characterizing systems along five core dimensions: environmental interaction sophistication, goal-directed behavior complexity, temporal coherence, learning and adaptation, and autonomy. This dimensional approach offers key advantages over existing definitions: it provides structured vocabulary that preserves nuance while adding precision, focuses on fundamental characteristics rather than specific implementations to accommodate evolving AI capabilities, and supports effective system evaluation by acknowledging the multifaceted nature of AI agents. By providing this structured framework for discussing and evaluating AI systems, we aim to facilitate more rigorous research methodology, enable clearer communication across AI subfields, and support the development of effective policies that can adapt to continued technological advancement.